\documentclass[conference]{IEEEtran}
\IEEEoverridecommandlockouts
\usepackage{cite}
\usepackage{amsmath,amssymb,amsfonts}
\usepackage{algorithmic}
\usepackage{graphicx}
\usepackage{textcomp}
\usepackage{float}
\usepackage{xcolor}
\usepackage{tikz}
\usepackage{subcaption}
\usepackage{multirow}
\usepackage{url}

\begin{document}

\date{}

\title{REVIVE: A Multi-Modal Framework for Vandalism Detection and Recovery in Autonomous Vehicles}

\author{
{\rm Abdullah Tariq Choudhry}\\
University of the Pacific\\
Stockton, CA, USA\\
a\_choudhry@u.pacific.edu
\and
{\rm Tapadhir Das}\\
University of the Pacific\\
Stockton, CA, USA\\
tdas@pacific.edu
} 

\maketitle

\begin{abstract}
Autonomous vehicles (AVs) face increasing threats from vandalism-induced occlusion attacks (VOAs) that compromise camera-based perception. While detection frameworks can identify vandalized images, restoring camera-stream utility after physical occlusion remains underexplored. This paper presents the Recovery and Enhancement of Vandalized Images for Vision Excellence (REVIVE) framework, a vandalism recovery pipeline integrating: (1) binary VOA detection, (2) multi-class VOA pattern identification, (3) EfficientNet-based U-Net segmentation, and (4) type-aware recovery using Bootstrapping Language-Image Pre-training (BLIP)-guided Stable Diffusion inpainting, direct pixel replacement, or adaptive median filtering. Stable Diffusion shows variable reconstruction performance (per-pattern SSIM 0.667-0.867, PSNR 15.4-26.7dB) across VOA patterns, while aligned direct pixel replacement achieves near-identical reconstruction (whole-image SSIM 0.988, PSNR 52.8dB; pixel-identical within the recovered region) under the aligned-reference condition. We emphasize that direct pixel replacement is a reference-based upper bound that presumes an available, spatially aligned clean reference frame, not a generally deployable recovery method. On 500 tracked clean/vandalized image pairs, unrecovered VOAs reduce YOLOv8l object-detection recall to 0.588, while direct pixel replacement restores recall to 0.967 and F1-score to 0.970 under that aligned-reference condition. LaMa, Telea, and Navier-Stokes baselines improve image similarity but provide more limited downstream detection recovery, and Stable Diffusion is treated as an asynchronous recovery branch subject to a quality gate rather than a blocking real-time perception step. We evaluate a reference-available quality gate that filters recovered candidates before downstream use: without it, type-aware routing degrades per-image recall to 0.304, whereas with it, recall returns to 0.608, at or above the unrecovered baseline, ensuring the forwarded stream is never worse than the unrecovered frame. REVIVE therefore provides a structured recovery framework and a practical evaluation of when each recovery strategy is appropriate for use.
\end{abstract}


\section{Introduction}
\label{sec:introduction}

\begin{figure}[t]
    \centering
    \includegraphics[width=\columnwidth]{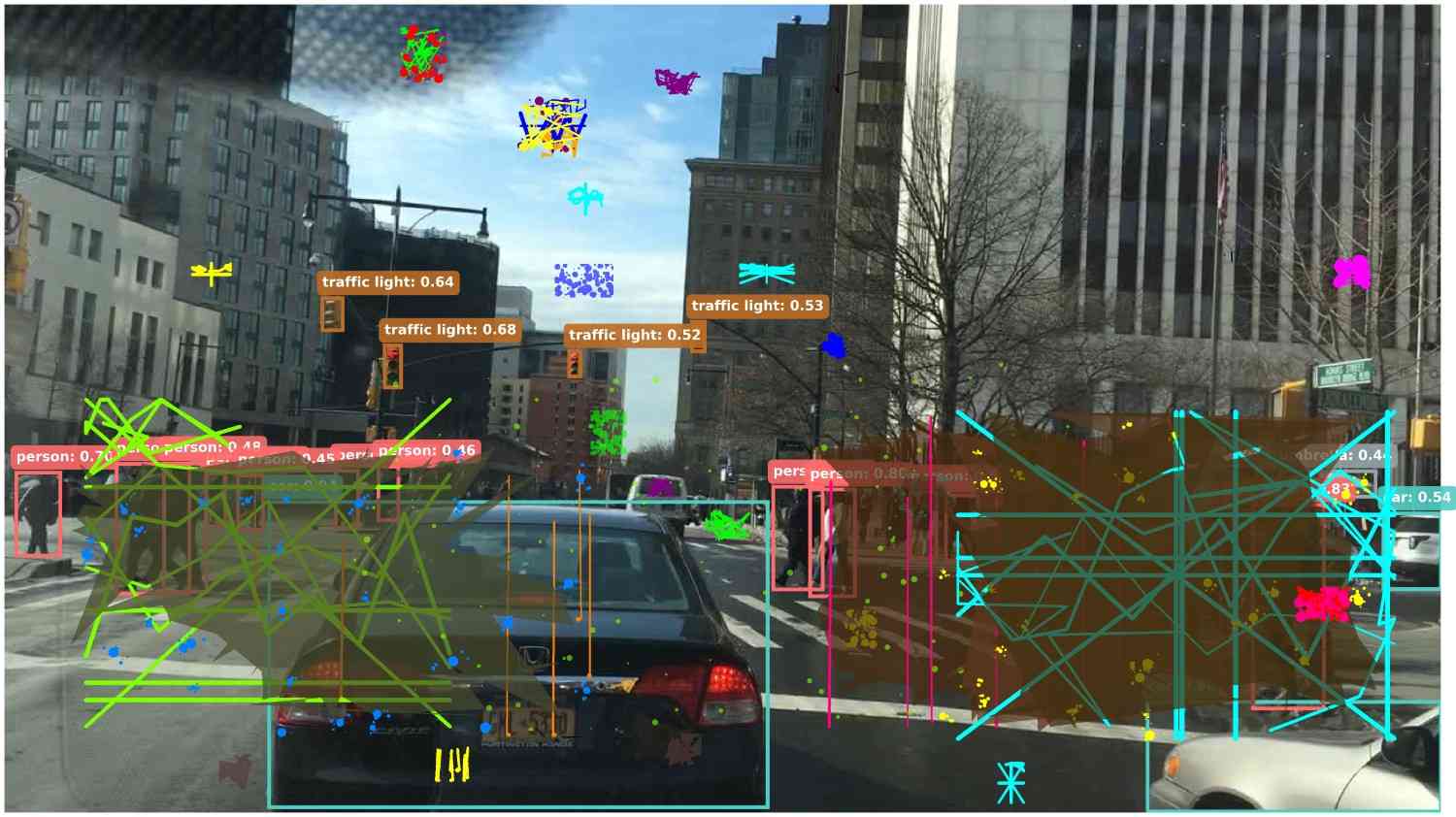}
    \caption{Example camera-frame degradation from a VOA, where physical occlusion blocks scene content before object detection or downstream planning can use the image.}
    \label{fig:intro_av_perception}
\end{figure}

The proliferation of autonomous vehicles (AVs) represents one of the most significant technological advances in modern transportation, with the global AV market projected to reach \$980.7 billion by 2040~\cite{alliedmarketresearch2025}. These systems rely heavily on camera-based sensors for real-time perception and decision-making. However, increasing AV deployment has exposed them to physical adversarial attacks targeting sensor systems through deliberate vandalism. Vandalism-induced occlusion attacks (VOAs) compromise camera functionality through spray paint, stickers, mud, debris, covers, or other physical obstructions~\cite{petit2015remote}. Unlike digital adversarial perturbations, physical vandalism can block or distort scene content directly and can persist until the camera is cleaned or replaced. Recent Waymo robotaxi vandalism incidents in San Francisco~\cite{verge2024waymofire, sfstandard2024waymotagging} highlight perception-system exposure to real physical abuse. An example of a VOA is illustrated in Figure~\ref{fig:intro_av_perception}.

Although modern AV platforms integrate multiple sensing modalities, including LiDAR, radar, and ultrasonic sensors, this work focuses specifically on camera-based perception for three reasons. First, cameras are the primary source of semantic information for object detection, lane recognition, and traffic sign interpretation; LiDAR and radar provide geometry and velocity but cannot replicate this semantic richness. Second, camera lenses are externally exposed and require no specialized knowledge or equipment to obstruct physically, making them the lowest-barrier attack surface. Third, VOA patterns such as spray paint, stickers, and mud are visually identifiable occlusion types that map directly onto image-domain classification and recovery techniques; sensor-level integrity monitoring for these modalities therefore naturally operates on the camera image stream. Our methods can complement broader multi-sensor fault-tolerance architectures, but the recovery problem addressed here is inherently image-level.

Modern approaches have attempted to address the susceptibility of AVs to VOAs~\cite{fayyad2020sensorfusion, matos2024survey}. However, these techniques restore functionality, primarily through sensor redundancy or fallbacks. This compromises the operational capability of the perception system; therefore, it continues to leave it vulnerable to prolonged disruption and safety hazards. Additionally, the critical challenge of restoring functionality through automated recovery remains unaddressed. To address this gap, this paper introduces the Recovery and Enhancement of Vandalized Images for Vision Excellence (REVIVE) framework, a vandalism recovery pipeline integrating detection, classification, localization, and restoration. Our four-stage framework employs: (1) binary classification (vandalized vs. clean), (2) multi-class VOA pattern identification, (3) EfficientNet-based U-Net segmentation for vandalism localization~\cite{tan2019efficientnet}, and (4) type-aware recovery using statistical filtering, direct pixel replacement, or Bootstrapping Language-Image Pre-training (BLIP)-guided Stable Diffusion reconstruction~\cite{li2022blip, rombach2022high}. The central novelty of REVIVE is its type-aware recovery strategy. Existing image inpainting and restoration methods treat all missing or corrupted regions uniformly, applying the same algorithm regardless of the structural nature of the corruption. This is suitable for general image completion tasks, but is suboptimal for VOA recovery, where the spatial structure of the occlusion varies significantly across different attack types. A scattered random occlusion is noise-like and responds well to statistical filtering; a large, coherent block obscuring the center of the image requires structured content synthesis. When an aligned, clean reference frame is available, direct pixel replacement achieves near-perfect fidelity at minimal cost. REVIVE couples multi-class VOA pattern identification with recovery strategy selection in an end-to-end pipeline, routing each occlusion type to the most appropriate recovery branch rather than applying a single method universally. This design directly addresses the quality-latency tradeoff: lightweight methods are reserved for cases where they are sufficient, and expensive generative methods are applied only where structurally necessary and treated as asynchronous when their runtime exceeds real-time constraints.

The main contributions of this work are:
\begin{itemize}
    \item Proposing REVIVE, an end-to-end framework that couples multi-class VOA classification with type-aware recovery branch selection for AV camera streams.
    \item Defining a threat model and deployment scope for image-level recovery within an AV perception stack. 
    \item Evaluating structured and unstructured VOAs using image quality metrics and aggregate downstream object detection recovery. 
    \item Comparing direct pixel replacement against classical and learned inpainting baselines and identifying when generative recovery is unsuitable for blocking real-time use. 
    \item Implementing and evaluating a reference-available quality gate that filters recovered candidates so the forwarded stream is never worse than the unrecovered frame for downstream detection.
\end{itemize}

The rest of the paper is as follows: Section~\ref{literature} reviews related work, Section~\ref{method} details our pipeline architecture, Section~\ref{results} presents experimental evaluation, and Section~\ref{conclusion} concludes with future directions.

\section{Literature Review}
\label{literature}

Physical adversarial attacks against AV perception systems emerged as a significant security concern following initial demonstrations by Szegedy et al.~\cite{szegedy2013intriguing} and Goodfellow et al.~\cite{goodfellow2014explaining}. Eykholt et al.~\cite{eykholt2018robust} showed physical adversarial examples remain effective under real-world variations in viewing angle, lighting, and resolution. Brown et al.~\cite{brown2017adversarial} introduced printable adversarial patches, while Chen et al.~\cite{chen2018shapeshifter} developed ShapeShifter attacks targeting AV object detection. Petit et al.~\cite{petit2015remote} demonstrated AV vulnerability to physical occlusion attacks, highlighting approaches that create realistic occlusions indistinguishable from environmental factors. Others have explored adversarial patches~\cite{thys2019fooling}, localized adversarial noise~\cite{karmon2018lavan}, and even physical accessories to fool perception systems~\cite{sharif2016accessorize}.

Non-restoration defense mechanisms have been proposed for adversarial attacks. Madry et al.~\cite{madry2017towards} introduced adversarial training through projected gradient descent, improving model robustness by training on adversarially perturbed examples. Papernot et al.~\cite{papernot2016distillation} demonstrated defensive distillation as a defense strategy, reducing model sensitivity to perturbations through knowledge distillation. Sensor fusion approaches have also been explored~\cite{fayyad2020sensorfusion}, with multiple works proposing redundant sensor systems combining camera, LiDAR, and radar to mitigate single-sensor failures. However, these approaches focus on digital perturbations rather than physical occlusions or rely on costly sensor redundancy that reduces operational efficiency.

To prevent the impact of physical occlusion attacks, many authors have proposed restoration methods~\cite{pathak2016context, yu2018generative, nazeri2019edgeconnect, suvorov2022resolution}. \cite{pathak2016context} introduced Context Encoders using CNNs with adversarial training to predict missing content. \cite{yu2018generative} advanced the field with Contextual Attention mechanisms borrowing information from distant spatial locations, while Nazeri et al.~\cite{nazeri2019edgeconnect} introduced EdgeConnect, leveraging edge information as structural guidance. Other notable works include using partial convolutions for irregular holes~\cite{liu2018image} and gated convolutions for free-form inpainting~\cite{yu2019free}. More recently, diffusion models have shown strong performance in image inpainting tasks~\cite{lugmayr2022repaint, saharia2022palette}. \cite{suvorov2022resolution} developed LaMa using Fourier convolutions for resolution-robust inpainting of large missing regions. While promising, these works focus primarily on general image completion. AV vandalism recovery differs because recovery must preserve perception-relevant objects, avoid hallucinating safety-critical content, and operate under latency constraints. REVIVE therefore evaluates restoration not only by image similarity, but also by downstream object-detection consistency using YOLOv8~\cite{jocher2023ultralytics}.

\section{Methodology}
\label{method}

\begin{figure*}[t]
    \centering
    \includegraphics[width=\textwidth]{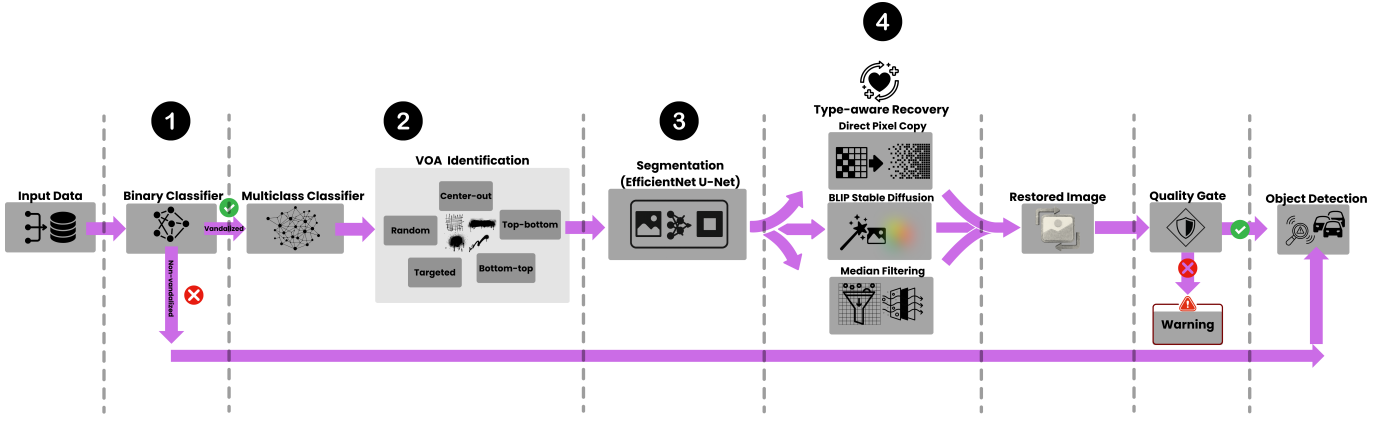}
    \caption{REVIVE pipeline architecture. Clean frames bypass recovery; vandalized frames are classified, segmented, and recovered by a type-aware module, then passed through a quality gate: frames that pass are forwarded to downstream object detection, while frames that fail are not output as restored imagery but instead raise a low-confidence (degraded-mode) warning.}
    \label{fig:pipeline_overview}
\end{figure*}

This section presents REVIVE, a vandalism recovery pipeline integrating detection, localization, and multi-modal restoration to address VOAs on AV camera systems. In this paper, recovery means improving the utility of an occluded camera frame for downstream perception, measured by image fidelity and object-detection consistency relative to the corresponding clean frame. It does not claim to prove full vehicle-level planning or control correctness. The framework employs a hierarchical strategy determining vandalism presence, identifying VOA type, localizing affected regions, applying appropriate recovery, and forwarding only outputs that pass the quality gate to downstream perception. The REVIVE framework is illustrated in Figure~\ref{fig:pipeline_overview}. The pipeline can be represented as:

\begin{equation}
\mathcal{P}(I) = \begin{cases}
\text{OD}(I) & \text{if } f_{\text{binary}}(I) = 0 \\
\text{OD}(R(I, S(I), f_{\text{multi}}(I))) & \text{if } f_{\text{binary}}(I) = 1
\end{cases}
\end{equation}

where $I$ is the input image, $f_{\text{binary}}$ and $f_{\text{multi}}$ are classification functions, $S$ is segmentation, $R$ is recovery, and $\text{OD}$ is object detection evaluation. This approach allows early termination for clean images and applies recovery only when the camera frame is likely compromised.

\subsection{Threat Model and Deployment Scope}
REVIVE assumes an attacker can physically occlude or contaminate one camera lens after deployment, but cannot modify the AV software stack, alter stored frames, or compromise other sensors. The attack objective is to reduce the utility of camera-based perception by hiding or distorting scene content. We focus on image-level recovery for a camera preprocessing module placed between sensor acquisition and object detection. In a practical AV stack, REVIVE would operate after camera capture and calibration but before the object detector, semantic perception module, sensor-fusion layer, planner, and controller. Its output is therefore a candidate perception input, not a direct driving command. If recovery confidence is low, the recovered frame should not be treated as authoritative; instead, the AV should fall back to existing safety mechanisms such as multi-sensor fusion, degraded-mode driving, or safe stop policies. This scope is intentionally narrower than full AV decision recovery: REVIVE evaluates whether a camera image can be restored enough to support object detection, not whether the vehicle can complete a maneuver safely.

\subsection{Stage 1: Binary Vandalism Detection}
The initial detection stage employs a convolutional neural network (CNN) architecture designed to distinguish between vandalized and non-vandalized automotive imagery. The main purpose of this stage is to ensure that the AV perception system is actively detecting when vandalism is occurring on the camera lens. The network utilizes a lightweight, three-block convolutional architecture with 3$\times$3 kernels, batch normalization, and ReLU activation after each convolutional layer, global average pooling, and a fully connected classification head with dropout regularization to prevent overfitting. This design prioritizes computational efficiency for rapid initial screening.

Input images are preprocessed through resizing to a standard resolution and normalized using ImageNet statistics to ensure consistency with pre-trained feature extractors. The model is trained using a Binary Cross-Entropy (BCE) loss function, defined as:
\begin{equation}
\mathcal{L}_{\text{BCE}} = - \frac{1}{N} \sum_{i=1}^{N} [y_i \log(\hat{y}_i) + (1 - y_i) \log(1 - \hat{y}_i)]
\end{equation}
where $N$ is the batch size, $y_i$ is the ground truth label (0 for non-vandalized, 1 for vandalized), and $\hat{y}_i$ is the predicted probability for the $i$-th image. The architecture employs global average pooling to reduce spatial dimensions while preserving feature information, followed by fully connected layers for final binary classification. This design choice provides computational efficiency while maintaining the spatial awareness necessary for robust vandalism detection across varied automotive scenes.

\subsection{Stage 2: Multi-Class VOA Classification}
\begin{figure}[t]
    \centering
    \begin{subfigure}[b]{0.3\columnwidth}
        \centering
        \includegraphics[width=\linewidth]{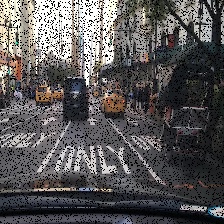}
        \caption{Random}
    \end{subfigure}
    \hfill
    \begin{subfigure}[b]{0.3\columnwidth}
        \centering
        \includegraphics[width=\linewidth]{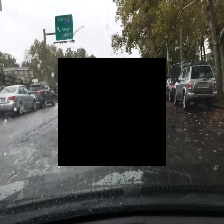}
        \caption{Center-out}
    \end{subfigure}
    \hfill
    \begin{subfigure}[b]{0.3\columnwidth}
        \centering
        \includegraphics[width=\linewidth]{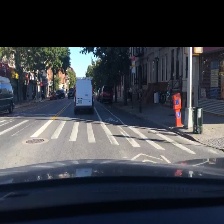}
        \caption{Top-bottom}
    \end{subfigure}
    
    \vspace{0.05cm}
    
    \begin{subfigure}[b]{0.3\columnwidth}
        \centering
        \includegraphics[width=\linewidth]{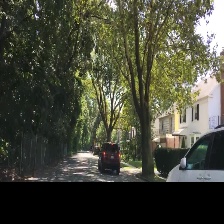}
        \caption{Bottom-top}
    \end{subfigure}
    \hfill
    \begin{subfigure}[b]{0.3\columnwidth}
        \centering
        \includegraphics[width=\linewidth]{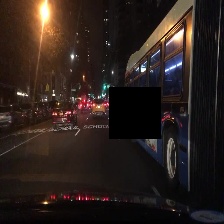}
        \caption{Targeted}
    \end{subfigure}
    
    \caption{Five VOA categories used in the framework.}
    \label{fig:vandalism_types}
\end{figure}

Upon vandalism detection, the system proceeds to identify the specific VOA pattern through multi-class classification. This stage selects the recovery strategy most appropriate for the spatial structure of the occlusion. We use five VOA categories based on their spatial characteristics and practical manifestations: \textbf{Random VOA} represents scattered occlusions that mimic salt-and-pepper noise, typically resulting from spray paint splatters or debris impacts. \textbf{Center-out VOA} involves square patches positioned at central locations, simulating targeted stickers or tape applications on camera sensors. \textbf{Top-bottom VOA} creates horizontal bands from the upper image regions, modeling scenarios where paint or covers are applied to the upper portion of cameras. \textbf{Bottom-top VOA} produces horizontal occlusions from lower regions, representing mud, dirt, or debris accumulation smeared on cameras. \textbf{Targeted VOA} places randomly positioned square patches within specific regions, simulating precise attacks on particular objects or areas of interest. An illustration of the different categories is shown in Figure~\ref{fig:vandalism_types}.

The multi-class classifier shares architectural similarities with the binary detector but is trained using a Categorical Cross-Entropy (CCE) loss function to output a probability distribution across the five VOA categories, defined by:
\begin{equation}
\mathcal{L}_{\text{CCE}} = - \frac{1}{N} \sum_{i=1}^{N} \sum_{c=1}^{C} y_{i,c} \log(\hat{y}_{i,c})
\end{equation}
where $C$ is the number of classes, $y_{i,c}$ is the one-hot encoded ground truth label, and $\hat{y}_{i,c}$ is the predicted probability for class $c$. This classification directly informs the subsequent restoration strategy selection, as different VOAs require distinct recovery approaches for optimal results.

\subsection{Stage 3: VOA Segmentation Localization}
Precise localization of vandalized regions is critical for effective restoration upon the identified VOA location. Our approach employs an EfficientNet-based U-Net architecture that combines the efficiency of EfficientNet backbones with the spatial precision of U-Net segmentation frameworks. This hybrid approach provides both computational efficiency and accurate pixel-level vandalism detection. The segmentation model incorporates VOA-type information from the previous stage as conditional input. This is achieved by transforming the class label into a feature vector that is then integrated into the segmentation network, allowing it to adapt its focus based on the expected spatial characteristics of a given VOA. For example, the class vector can be concatenated with the encoder's feature map before the decoder stage. This conditioning mechanism improves segmentation accuracy by leveraging prior knowledge about expected spatial patterns for different VOAs.

The U-Net decoder reconstructs spatial resolution through transposed convolutions and skip connections, producing binary masks that precisely delineate vandalized regions. The model is trained using a combined loss function that balances pixel-wise accuracy with region-based similarity, particularly effective for segmentation tasks. This loss is a weighted sum of Binary Cross-Entropy ($\mathcal{L}_{\text{BCE}}$) and a Dice Loss ($\mathcal{L}_{\text{Dice}}$):
\begin{equation}
\mathcal{L}_{\text{seg}} = \alpha \mathcal{L}_{\text{BCE}} + \beta \mathcal{L}_{\text{Dice}}
\end{equation}
where $\alpha$ and $\beta$ are weighting factors to balance the two terms. The Dice Loss is defined as:
\begin{equation}
\mathcal{L}_{\text{Dice}} = 1 - \frac{2 \sum_{i=1}^{N} p_i g_i}{\sum_{i=1}^{N} p_i^2 + \sum_{i=1}^{N} g_i^2}
\end{equation}
where $p_i$ are the predicted pixel probabilities and $g_i$ are the ground truth pixel labels. These masks serve as input to the subsequent restoration stage, enabling targeted recovery while preserving unaffected image content.

\subsection{Stage 4: Type-Aware Image Recovery}
The restoration stage employs multiple recovery strategies optimized for different VOAs and available resources. Our framework implements three distinct approaches: statistical-parametric median filtering for random occlusions, BLIP-guided Stable Diffusion reconstruction for complex structured patterns, and direct pixel replacement when aligned reference imagery is available. The selection of a recovery method is contingent on the VOA type. Statistical median filtering is applied to random VOAs because of their scattered, noise-like occlusions, which can be reduced without semantic scene synthesis. For structured VOAs (e.g., center-out, top-bottom), which involve large, coherent occluded regions, generative inpainting can synthesize semantically plausible content but must be treated cautiously because it can hallucinate safety-relevant objects. Finally, direct pixel replacement is evaluated as a reference-frame upper bound: it is appropriate only when a recent clean frame is available and remains spatially aligned with the current view. This assumption is plausible for short temporal windows in stopped or slow-moving conditions, but it degrades as ego-motion, object motion, or lighting changes increase.

\subsubsection{Statistical-Parametric Median Filtering}
For random VOAs, we develop a sophisticated median filtering approach that dynamically adapts parameters based on image characteristics. The method employs histogram-derived adaptive thresholding using $(\mu - 2\sigma)$ clamped bounds [10,50], multi-scale morphological clustering analysis across kernel sizes [3×3, 5×5, 7×7, 9×9], and conservative kernel selection with quality-first philosophy (bias toward 3×3, maximum 5×5). The adaptive dark threshold is calculated as:
\begin{equation}
T_{\text{adaptive}} = \max(10, \min(50, \mu_{\text{noise}} - 2\sigma_{\text{noise}}))
\end{equation}

Maximum processing rounds are determined using complexity-weighted logarithmic scaling:
\begin{equation}
\begin{split}
R_{\max} = \max(1, \min(&\lfloor\log_{10}(N + 1) \times C \times S\rfloor, \\
                     &\max(3, \lfloor\log_2(N + 1)\rfloor)))
\end{split}
\end{equation}

where $N$ is noise count, $C$ is complexity multiplier, and $S$ is size factor. Convergence detection uses multi-criteria analysis, including improvement trend analysis, plateau detection, and diminishing returns patterns.

\subsubsection{BLIP-Guided Stable Diffusion Reconstruction}
Structured VOAs can benefit from generative approaches that synthesize semantically plausible content for large occluded regions. Our BLIP-guided Stable Diffusion pipeline integrates BLIP vision-language understanding with Stable Diffusion inpainting to create context-aware reconstructions. The process begins with scene analysis using BLIP to generate descriptive prompts from available scene context. These prompts guide Stable Diffusion inpainting over the segmented vandalized region. Because diffusion-based reconstruction is stochastic and can introduce visual artifacts or hallucinated objects, REVIVE treats this branch as an asynchronous semantic-recovery option subject to a quality gate rather than a blocking replacement for raw camera frames in the real-time perception loop.

\subsubsection{Direct Pixel Replacement}
When an aligned clean reference frame is available, direct pixel replacement provides an upper-bound reconstruction with minimal error. This is defined by:

\begin{equation}
I_{\text{recovered}}(x,y) = \begin{cases}
I_{\text{ref}}(x,y) & \text{if } M(x,y) = 1 \\
I_v(x,y) & \text{if } M(x,y) = 0
\end{cases}
\end{equation}

where $M(x,y)$ is the binary segmentation mask indicating vandalized pixels, $I_{\text{ref}}$ is the aligned reference image, and $I_v$ is the vandalized image. This is not a general solution for all driving conditions; direct pixel replacement is an upper-bound reference method for short-horizon temporal recovery when the reference frame has not aged significantly. To stress this assumption, we also tested deliberately mismatched clean references from other tracked samples of the same VOA type. This proxy for reference aging reduced average SSIM to 0.720 and average detection recall to 0.586, confirming that direct pixel replacement should be accepted only when temporal alignment is verified.

\subsection{Quality Gate for Recovery Outputs} \label{sec:quality_gate}
To reduce the risk of low-confidence or inconsistent reconstructions entering the perception stack, REVIVE applies a post-recovery quality gate. Given a recovered candidate frame, the gate compares it against the detector's clean-frame behavior and against simply leaving the frame unrecovered, and accepts the candidate only if it (i) does not suppress safety-relevant detections (recall no lower than the unrecovered frame), (ii) does not introduce excessive new detections (no increase in false positives), and (iii) clears a structural-consistency floor (SSIM $\geq 0.5$). If a candidate passes all three criteria, the recovered frame is forwarded to downstream perception as the active camera input for that frame. If it fails any criterion, the candidate is rejected: REVIVE does not output a restored frame and instead raises a low-confidence (degraded-mode) signal for that frame. How the downstream stack responds to this signal---increasing reliance on other sensors, issuing an operator warning, or invoking a cautionary driving policy---is a vehicle-level decision outside the scope of this paper; the essential property is that unverified reconstructed imagery is never presented to perception as reliable. We evaluate this reference-available gate on the tracked set in Section~\ref{sec:quality_gate_eval}. In deployment without ground-truth clean frames, the same criteria can be approximated using temporal consistency checks, multi-sensor disagreement, confidence drops, and object-level plausibility tests; this reference-free instantiation remains future work. Thus, REVIVE's operational contribution is not that every recovery branch is always better than every baseline; it is that recovery candidates are selected by VOA type and filtered before downstream use so that the forwarded stream is never worse than the unrecovered frame.

\subsection{Object Detection Evaluation}
The final pipeline stage evaluates recovery using object detection on restored images. We employ YOLOv8~\cite{jocher2023ultralytics} as a representative modern object detector. Detections on recovered or vandalized images are matched against detections on the paired clean frame using class agreement and bounding-box IoU. This protocol measures whether recovery restores perception-relevant objects, not merely whether reconstructed pixels look plausible.

\subsection{Quality Assessment Framework}
Throughout the pipeline, we employ comprehensive quality assessment metrics to evaluate recovery effectiveness. Structural Similarity Index Measure (SSIM)~\cite{wang2004image} and Peak Signal-to-Noise Ratio (PSNR) provide quantitative measures of restoration quality from the output provided by the REVIVE framework. At the same time, Intersection over Union (IoU), a common metric in object detection challenges like PASCAL VOC~\cite{everingham2010pascal}, assesses the segmentation performance of the proposed framework. PSNR is derived from the Mean Squared Error (MSE), defined as:
\begin{equation}
\text{MSE} = \frac{1}{mn} \sum_{i=0}^{m-1} \sum_{j=0}^{n-1} [I(i,j) - K(i,j)]^2
\end{equation}
where $I$ is the original and $K$ is the recovered image, both of size $m \times n$. PSNR is then calculated as:
\begin{equation}
\text{PSNR} = 10 \cdot \log_{10}\left(\frac{\text{MAX}_I^2}{\text{MSE}}\right)
\end{equation}
where $\text{MAX}_I$ is the maximum possible pixel value of the image. The IoU for segmentation is defined as:
\begin{equation}
\text{IoU} = \frac{|A \cap B|}{|A \cup B|} = \frac{\text{TP}}{\text{TP} + \text{FP} + \text{FN}}
\end{equation}

where $A$ is the predicted mask, $B$ is the ground truth mask, and TP, FP, and FN represent true positives, false positives, and false negatives, respectively. SSIM quantifies structural similarity between original and recovered images on a scale from 0 to 1, where higher values indicate better preservation of image structure. PSNR measures reconstruction fidelity in decibels, with higher values representing less distortion and better quality recovery. The quality assessment framework enables systematic comparison between different recovery approaches and provides guidance for optimal strategy selection based on specific application requirements and resource constraints.

\section{Experimentation Details and Results}
\label{results}

\subsection{Experimental Setup}
We use the BDD100K dataset~\cite{yu2020bdd100k}, a large-scale and diverse driving dataset. While other excellent datasets like KITTI~\cite{geiger2012we} and nuScenes~\cite{caesar2020nuscenes} exist, BDD100K's diversity in weather and scene types is particularly suitable for our vandalism study. We extract 5,000 images with vandalized versions across five VOA patterns. Each type includes 1,000 samples with 10-30\% occlusion. Training employs a 70-15-15 split. For downstream object detection, we additionally use 500 tracked clean/vandalized image pairs with 100 examples per VOA type. Each tracked pair preserves the clean reference, vandalized frame, generated mask, and recovery outputs, allowing aggregate comparison against clean-frame YOLOv8l detections. Performance metrics include accuracy, precision, recall, F1, IoU, Dice coefficient, SSIM, PSNR, processing time, and downstream detection precision/recall/F1. All reported latencies were profiled on Apple Silicon using PyTorch MPS.

\subsection{Experimental Results}
\begin{figure}[t]
    \centering
    \includegraphics[width=\columnwidth]{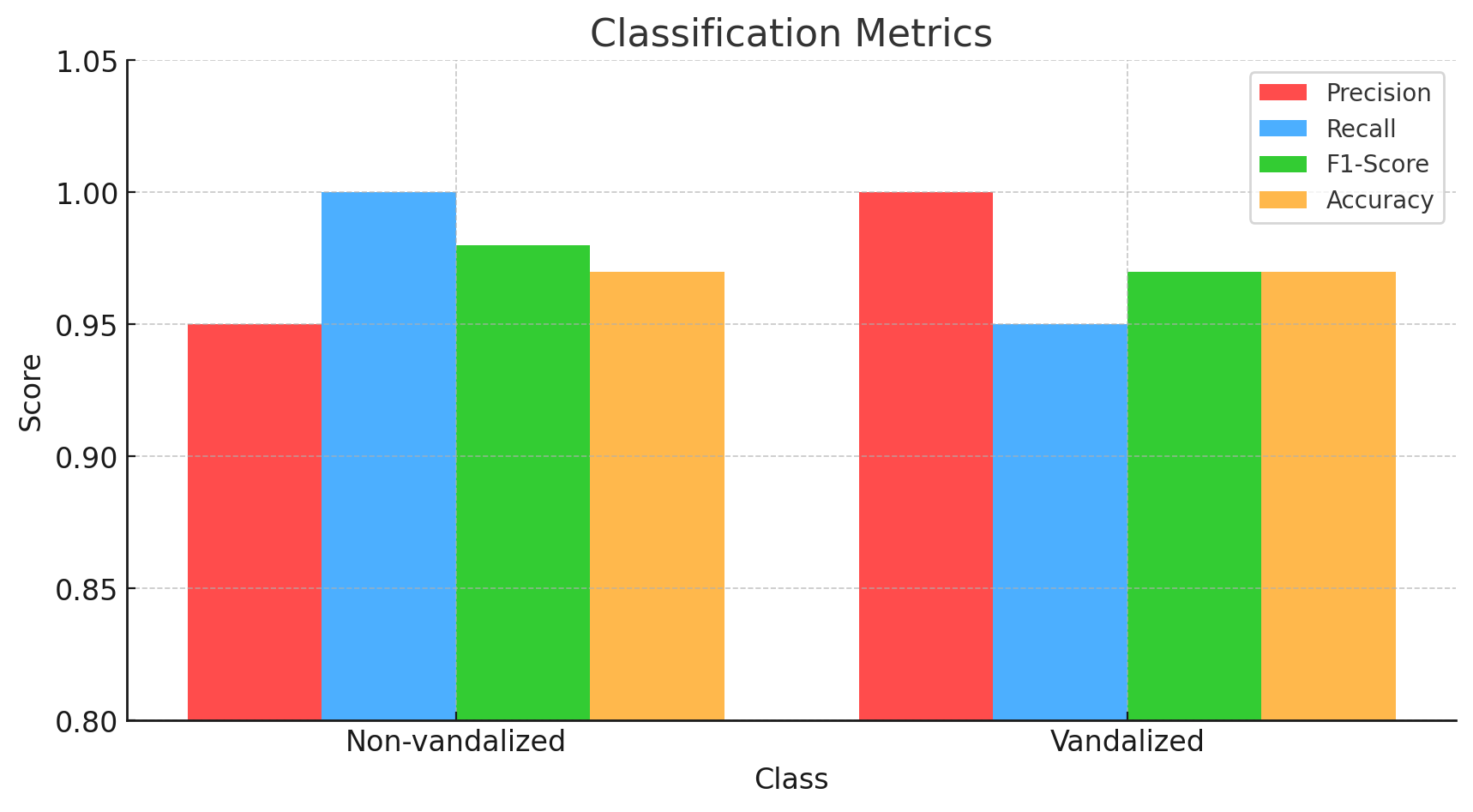}
    \caption{Binary vandalism detector performance}
    \label{fig:binary_classifier}
\end{figure}

First, we observe the results achieved by our binary vandalism detector, as shown in Figure~\ref{fig:binary_classifier}. We note that it achieves 97\% overall accuracy with precision and recall scores exceeding 95\% for both classes, ensuring reliable pipeline initialization for actively detecting vandalism.

\begin{figure}[t]
    \centering
    \includegraphics[width=\columnwidth]{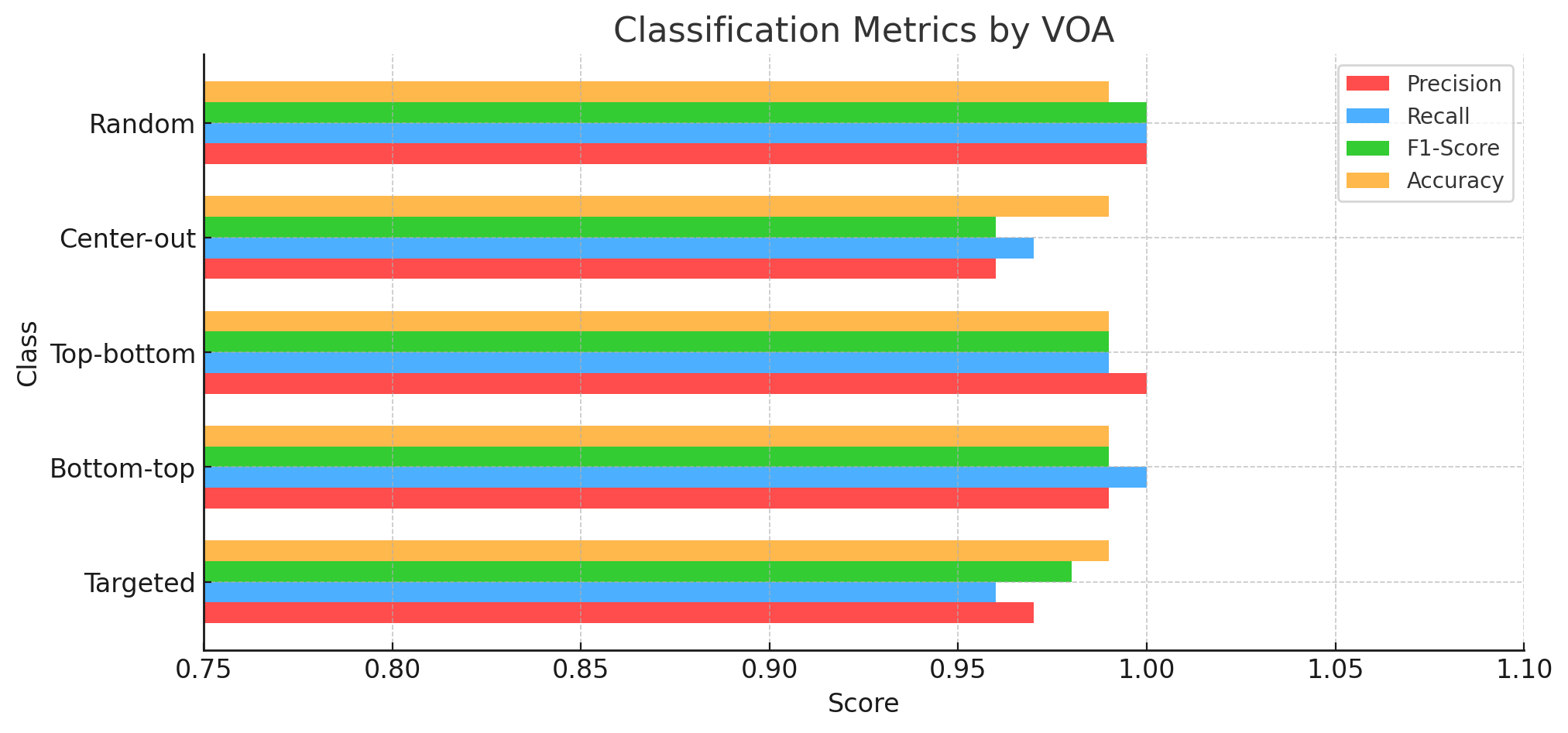}
    \caption{Multi-class VOA detector performance}
    \label{fig:multiclass_classifier}
\end{figure}

Next, we observe the performance achieved by the multi-class vandalism classifier, with results shown in Figure~\ref{fig:multiclass_classifier}. We note that the classifier demonstrates superior performance with 99\% overall accuracy across all five VOA types. Detailed per-class analysis shows consistent performance: random VOAs achieve 100\% precision and recall, center-out patterns achieve 96\% precision and 97\% recall, top-bottom attacks achieve 100\% precision and 99\% recall, bottom-top vandalism achieves 99\% precision and 100\% recall, and targeted attacks achieve 99\% precision and 97\% recall. This classification accuracy enables precise recovery strategy selection further down the REVIVE pipeline.

\begin{figure}[t]
    \centering
    \includegraphics[width=\columnwidth]{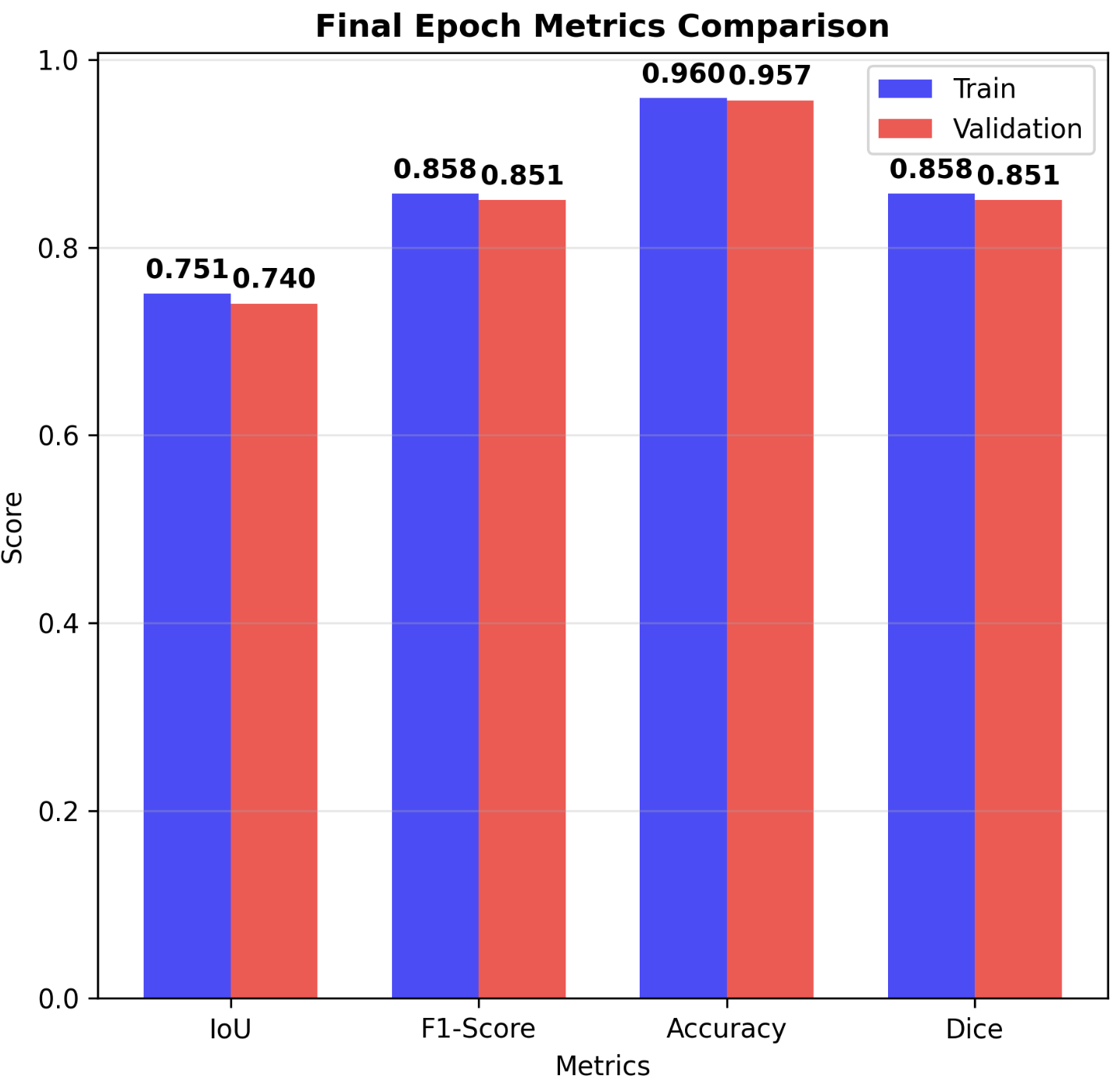}
    \caption{Final epoch performance of EfficientNet-U-Net}
    \label{fig:training_curves}
\end{figure}

\begin{figure}[t]
    \centering
    \begin{subfigure}{0.5\textwidth}
        \includegraphics[width=\textwidth]{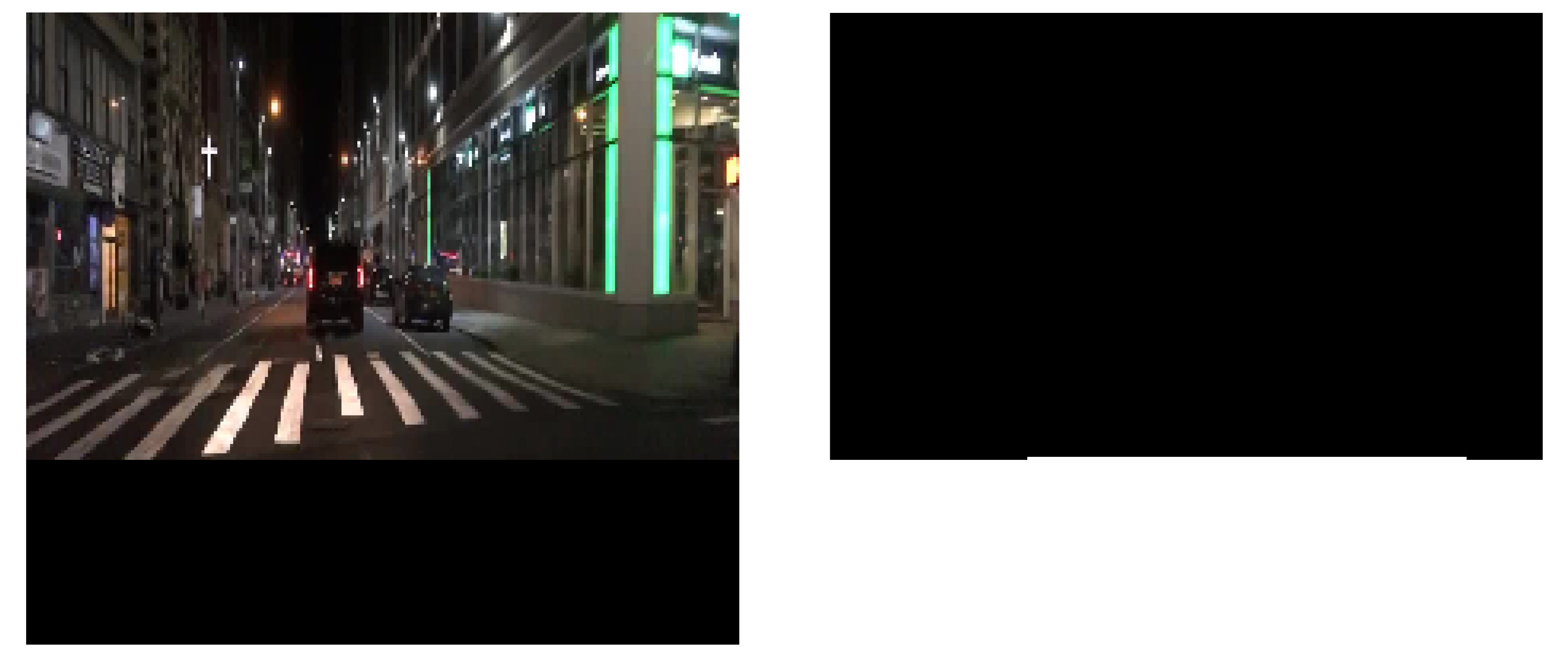}
        \caption{Bottom-top (IoU: 0.992)}
        \label{fig:iou99}
    \end{subfigure}
    \hfill
    \begin{subfigure}{0.5\textwidth}
        \includegraphics[width=\textwidth]{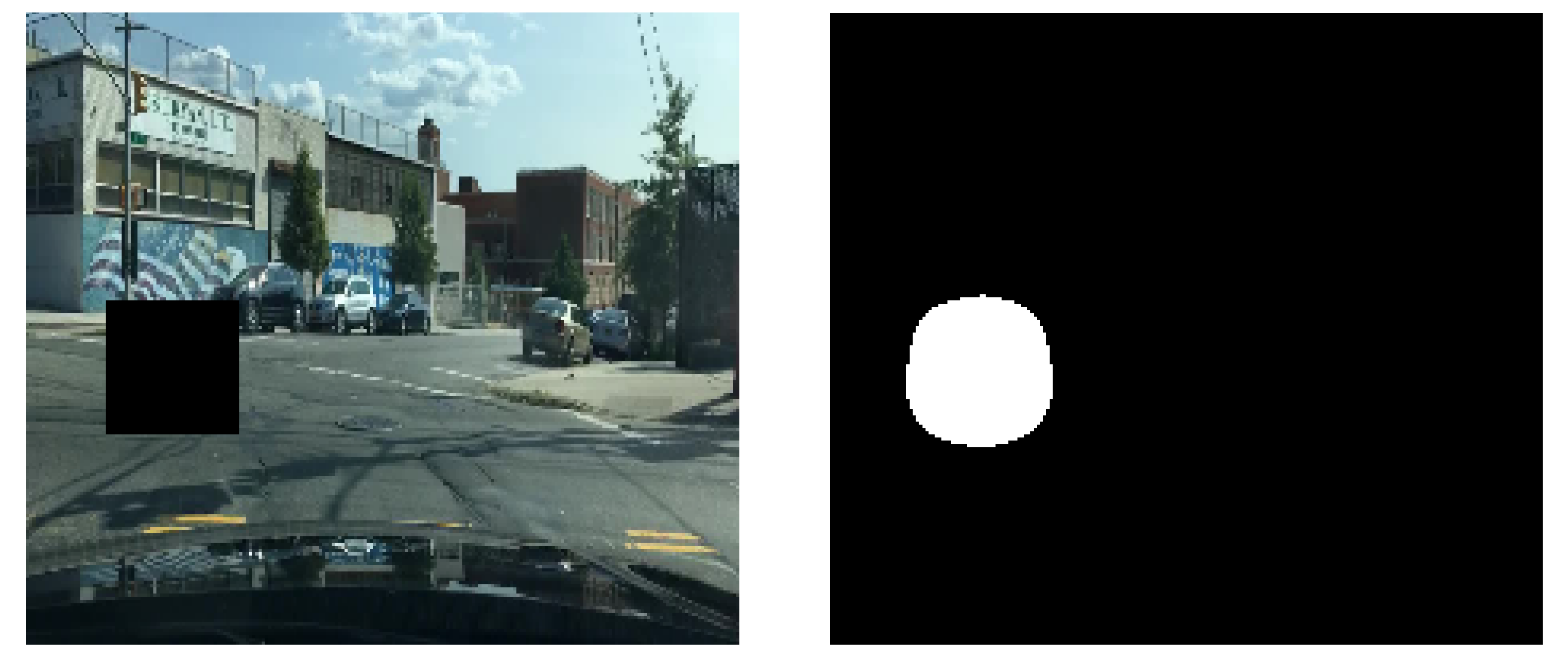}
        \caption{Targeted (IoU: 0.852)} 
        \label{fig:iou85}
    \end{subfigure}
    \caption{Segmentation results showing vandalized images (left) and predicted masks (right) for two VOA types.}
    \label{fig:segmentation_examples}
\end{figure}

Following detection and classification, the EfficientNet-based U-Net model achieves strong performance as shown in Figure~\ref{fig:training_curves} and Figure~\ref{fig:segmentation_examples}. 
In Figure~\ref{fig:training_curves}, we note that the Dice coefficient analysis confirms segmentation quality, with mean scores of 0.91 for structured VOAs and 0.87 for random vandalism. The conditional architecture incorporating VOA-type information demonstrates improved segmentation performance compared to unconditional baselines, validating the effectiveness of our type-aware approach. These results are further reinforced in Figure~\ref{fig:segmentation_examples}, where we observe the segmentation model achieving remarkable localization precision for Bottom-top (Figure~\ref{fig:iou99}) and Targeted VOAs (Figure~\ref{fig:iou85}), with mean Intersection over Union (IoU) scores exceeding 85\% across all VOA types. Structured VOAs (center-out, top-bottom, bottom-top, targeted) consistently achieve IoU scores above 88\%, while random VOAs achieve 82\% IoU due to their inherently scattered nature. The usage of the EfficientNet-based U-Net demonstrates suitable performance for real-time applications.

\begin{table}[t]
\centering
\small
\setlength{\tabcolsep}{3pt}
    \caption{Recovery performance summary. Unless marked otherwise, SSIM, PSNR, F1, recall, and clean-reference mAP50 are from the 500-image tracked evaluation. Pixel replacement is a reference-frame upper bound, not a general deployment guarantee.}
\label{table:recovery_comparison}
\resizebox{\columnwidth}{!}{%
\begin{tabular}{|l|c|c|c|c|c|c|}
\hline
\textbf{Method} & \textbf{SSIM} & \textbf{PSNR} & \textbf{F1} & \textbf{Recall} & \textbf{mAP50} & \textbf{Time} \\
\hline
Unrecovered VOA & 0.727 & 20.09dB & 0.815 & 0.588 & 0.635 & -- \\
Stable Diffusion$^{*}$ & 0.731 & 18.90dB & 0.492 & 0.339 & 0.162 & 8610ms \\
LaMa Inpaint & 0.845 & 26.42dB & 0.820 & 0.667 & 0.680 & 170ms \\
Pixel Replace & 0.988 & 52.82dB & 0.970 & 0.967 & 0.979 & 0.76ms \\
Telea Inpaint & 0.849 & 26.03dB & 0.818 & 0.607 & 0.622 & 8.48ms \\
Navier-Stokes & 0.848 & 25.95dB & 0.810 & 0.606 & 0.625 & 8.42ms \\
Median Filter & 0.660 & 19.22dB & 0.448 & 0.214 & 0.110 & 0.26ms \\
\hline
\end{tabular}%
}
\vspace{0.1cm}
\scriptsize
\textbf{Note:} Recovery times are in ms. mAP50 uses clean-frame YOLOv8l detections as the reference and matches recovered-frame detections by class and IoU $\geq$ 0.50. $^{*}$Stable Diffusion is evaluated on the 400 structured VOA samples where the generative branch applies; random VOAs use median filtering in REVIVE.
\end{table}

\begin{table}[t]
\centering
\small
\setlength{\tabcolsep}{3pt}
    \caption{Deployment-oriented latency budget for REVIVE recovery branches (all times in ms). Fast branches are candidates for the online perception path after quality gating; Stable Diffusion is reserved for asynchronous recovery or situations where the AV can tolerate delayed refinement.}
\label{table:latency_budget}
\resizebox{\columnwidth}{!}{%
\begin{tabular}{|l|c|l|}
\hline
\textbf{Component / Branch} & \textbf{Measured Time} & \textbf{Deployment Role} \\
\hline
Direct Pixel Replacement & 0.76ms & Online candidate if aligned reference exists \\
Median Filter & 0.26ms & Online candidate after quality gate \\
Telea / Navier-Stokes & 8.4ms & Near-online candidate after quality gate \\
LaMa Inpainting & 170ms & Learned baseline; not a blocking 30 FPS branch \\
Stable Diffusion & $\sim$8610ms & Asynchronous or pause-tolerant recovery branch \\
\hline
\end{tabular}%
}
\end{table}

\begin{table}[t]
\centering
\scriptsize
\setlength{\tabcolsep}{3pt}
    \caption{Direct pixel replacement alignment sensitivity on 500 tracked pairs. Shift is applied to the clean reference before replacement to approximate temporal misalignment/reference aging.}
\label{table:alignment_sensitivity}
\begin{tabular}{|c|c|c|c|}
\hline
\textbf{Reference Shift} & \textbf{SSIM} & \textbf{F1} & \textbf{Recall} \\
\hline
0 px & 0.988 & 0.970 & 0.967 \\
2 px & 0.856 & 0.857 & 0.802 \\
5 px & 0.814 & 0.818 & 0.665 \\
10 px & 0.795 & 0.802 & 0.618 \\
20 px & 0.775 & 0.794 & 0.595 \\
40 px & 0.754 & 0.776 & 0.578 \\
\hline
\end{tabular}
\end{table}

\begin{figure*}[t]
    \centering
    \includegraphics[width=\textwidth]{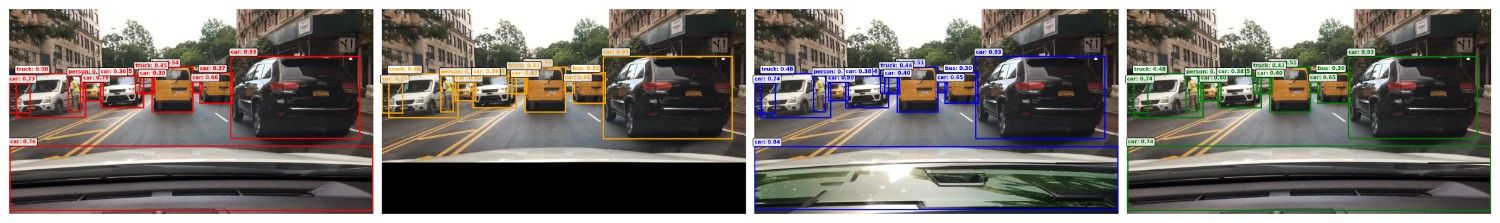}
    \caption{Representative bottom-top VOA recovery example showing clean reference, vandalized input, generative recovery, and direct pixel replacement. The figure illustrates qualitative branch behavior; aggregate detection metrics are reported separately in Tables~\ref{table:recovery_comparison} and~\ref{table:detection_stats}.}
    \label{fig:recovery_comparison_bottomtop}
\end{figure*}

\begin{figure*}[t]
    \centering
    \begin{subfigure}{0.48\textwidth}
        \includegraphics[width=\textwidth]{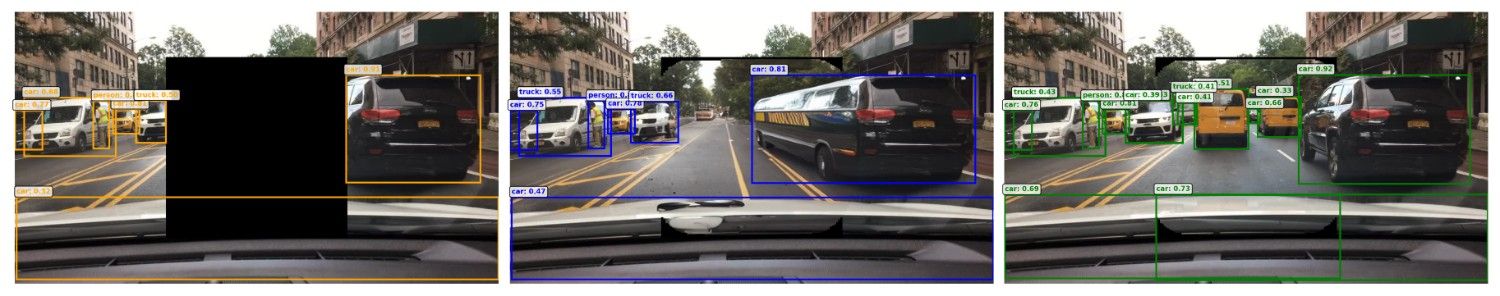}
        \caption{Center-out VOA recovery (Vandalized $\rightarrow$ GenAI $\rightarrow$ Pixel Copy)}
        \label{fig:recovery_comparison_centerout}
    \end{subfigure}
    \hfill
    \begin{subfigure}{0.48\textwidth}
        \includegraphics[width=\textwidth]{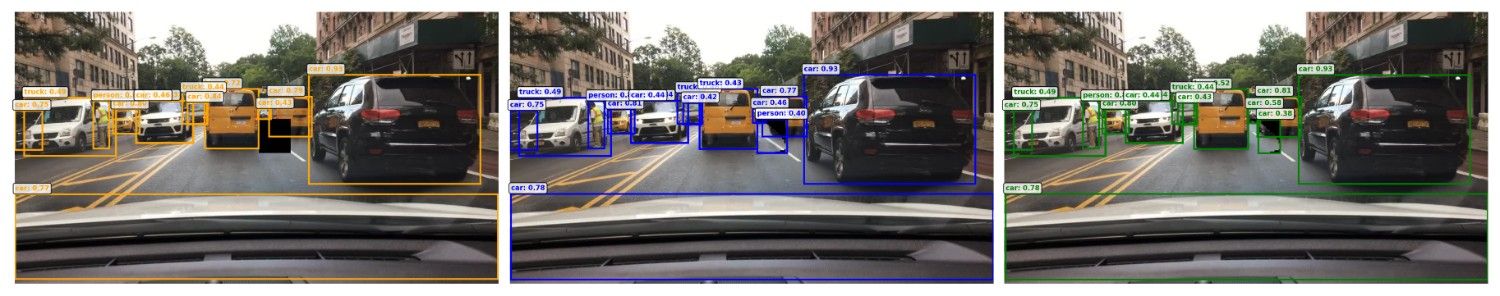}
        \caption{Targeted VOA recovery (Vandalized $\rightarrow$ GenAI $\rightarrow$ Pixel Copy)}
        \label{fig:recovery_comparison_targeted}
    \end{subfigure}
    \caption{Representative recovery examples for center-out and targeted VOAs. Center-out attacks obscure semantically important image regions and expose the reliability gap between plausible-looking generative outputs and reference-based recovery.}
    \label{fig:centertarget}
\end{figure*}

\begin{figure*}[t]
    \centering
    \begin{subfigure}{0.48\textwidth}
        \includegraphics[width=\textwidth]{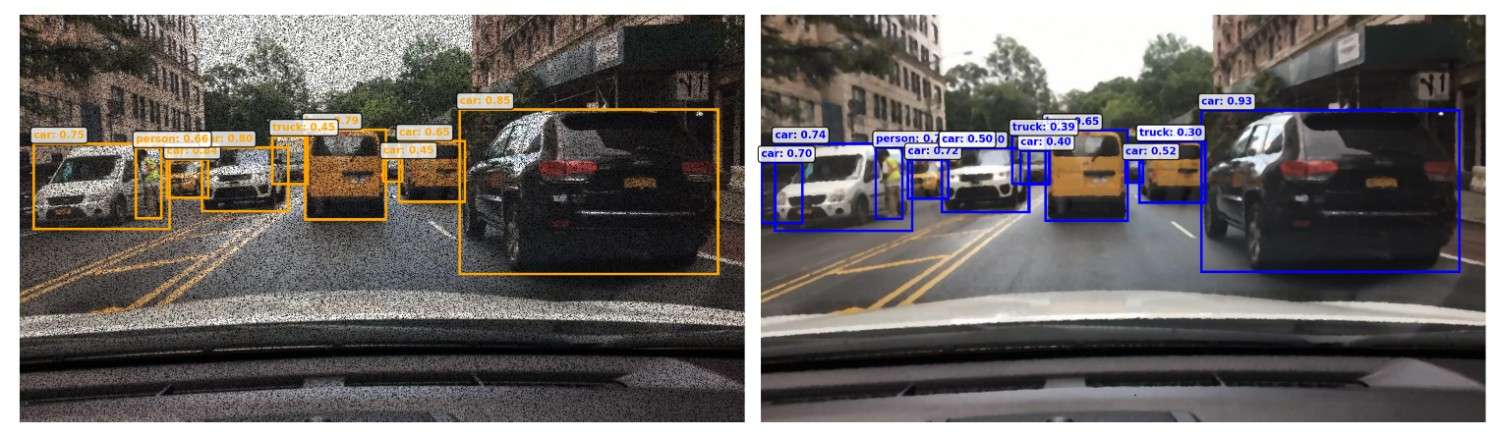}
        \caption{Random VOA recovery (Vandalized $\rightarrow$ Median Filtering)}
        \label{fig:recovery_comparison_random}
    \end{subfigure}
    \hfill
    \begin{subfigure}{0.48\textwidth}
        \includegraphics[width=\textwidth]{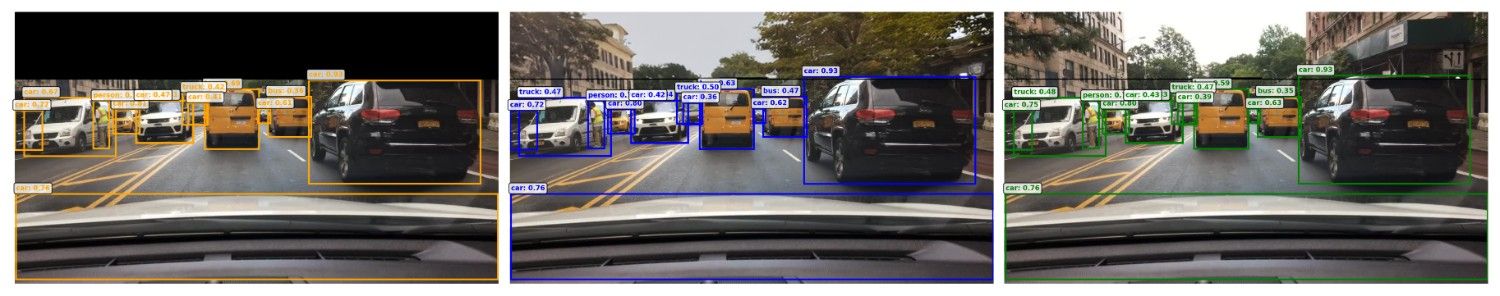}
        \caption{Top-bottom VOA recovery (Vandalized $\rightarrow$ GenAI $\rightarrow$ Pixel Copy)}
        \label{fig:recovery_comparison_topbottom}
    \end{subfigure}
    \caption{Representative recovery examples for random and top-bottom VOAs. Random attacks are treated as noise-like corruption, while top-bottom attacks require structured-region restoration or direct pixel replacement when an aligned clean reference is available.}
    \label{fig:randomtop}
\end{figure*}

With the vandalized regions localized, we now turn to evaluating the quality of our different recovery approaches across the five VOA types. Our analysis examines how each VOA pattern responds to the recovery methods employed by the REVIVE framework. Visual comparisons are presented in Figures~\ref{fig:recovery_comparison_bottomtop},~\ref{fig:centertarget},~\ref{fig:randomtop}, and quantitative trade-offs are summarized in Tables~\ref{table:recovery_comparison},~\ref{table:latency_budget},~\ref{table:e2e_latency}, and~\ref{table:alignment_sensitivity}.

\textbf{Bottom-top VOA Recovery:} Bottom-top VOAs create horizontal occlusions from lower image regions, often hiding road surface, lane context, or nearby objects. The qualitative examples show that generative recovery can preserve broad scene structure, but it may introduce artifacts in the reconstructed lower region. Direct pixel replacement is substantially more reliable when the reference remains aligned, but this result should be interpreted as an upper bound rather than a guarantee under arbitrary ego-motion.

\textbf{Center-out VOA Recovery:} Center-out patterns present significant challenges because they obscure central semantic content. This is the setting where an incorrect reconstruction is most dangerous: a plausible hallucination can look visually coherent while still failing to restore the object-detection evidence needed by an AV perception stack. REVIVE therefore routes such outputs through the quality gate and favors direct pixel replacement only when temporal alignment is strong.

\textbf{Targeted VOA Recovery:} Targeted VOAs affect localized regions and may leave enough surrounding context for plausible inpainting. Even in this favorable case, downstream detection consistency remains the key safety metric because visually realistic reconstruction does not necessarily imply object-level correctness.

\textbf{Random VOA Recovery:} Random VOAs are scattered, noise-like corruptions. Median filtering is fast and can visually suppress isolated artifacts, but the aggregate object-detection evaluation shows that it does not reliably restore downstream detections across all random samples. This reconciles the low SSIM/PSNR values with the qualitative appearance: the filter may reduce visible noise while also blurring or altering object cues needed by YOLOv8l.

\textbf{Top-bottom VOA Recovery:} Top-bottom VOAs create horizontal occlusions from upper image regions, modeling scenarios where paint, covers, or debris affect the upper camera view. These attacks may obscure traffic lights, signs, sky-line context, or distant vehicles. Generative recovery can synthesize plausible upper-scene content, but deployment must account for its latency and hallucination risk.

\subsection{Object Detection Performance Evaluation}
Finally, we evaluate the practical impact of our recovery methods on a downstream perception task. The results from YOLOv8 object detection on the restored imagery provide a clear measure of functional improvement.

\subsubsection{Comparative Analysis of Recovery Approaches}
To provide a comprehensive overview of the trade-offs between recovery strategies, Table~\ref{table:recovery_comparison} reports aggregate performance across 500 tracked clean/vandalized pairs, including clean-reference mAP50 computed against clean-frame YOLOv8l detections. Unrecovered VOAs reduce detection recall to 0.588 and mAP50 to 0.635. Direct pixel replacement restores recall to 0.967, F1-score to 0.970, and mAP50 to 0.979 in 0.76ms on average, confirming its value as an upper bound when the reference frame is valid. LaMa provides a learned inpainting baseline with SSIM 0.845, PSNR 26.42dB, recall 0.667, and mAP50 0.680, outperforming unrecovered frames and the classical baselines in recall while remaining substantially slower than the lightweight online candidates. Stable Diffusion provides semantically plausible structured-region recovery, but its aggregate structured-VOA detector recovery is lower (F1 0.492, recall 0.339, mAP50 0.162), and its 8.61s average runtime places it outside the blocking perception path. This result supports the REVIVE design choice: generative recovery is valuable as a quality-gated semantic branch, while detector-critical recovery should prefer faster reference-based or learned inpainting candidates when they pass object-level checks. Classical Telea and Navier-Stokes inpainting improve SSIM and PSNR over unrecovered frames but provide only modest detection-recall gains, suggesting that visual similarity alone is insufficient for AV perception recovery. Median filtering is fast, but its average detection recall of 0.214 shows that it should be restricted to cases where the quality gate indicates object-level improvement. Table~\ref{table:latency_budget} further separates online candidates from the asynchronous Stable Diffusion branch, addressing the frame-rate gap without discarding its reconstruction benefits.

Table~\ref{table:alignment_sensitivity} addresses the direct-pixel-replacement reference assumption. Under the aligned-reference condition, detection recall reaches 0.967. A controlled 2-pixel shift reduces recall to 0.802, and 5--20 pixel shifts reduce recall to 0.665--0.595. This confirms that direct pixel replacement is useful only when the reference frame is temporally and spatially aligned; otherwise, the recovered frame should not be accepted without additional consistency evidence.

\begin{table}[t]
\centering
\small
\setlength{\tabcolsep}{4pt}
\caption{Measured end-to-end latency of the online (non-generative) REVIVE path on Apple Silicon (MPS), averaged over 482 vandalized tracked frames. REVIVE preprocessing (22.7\,ms) fits within a 33\,ms (30\,FPS) budget; the end-to-end total includes YOLOv8l object detection, which an AV executes regardless of REVIVE.}
\label{table:e2e_latency}
\begin{tabular}{lc}
\hline
\textbf{Pipeline stage} & \textbf{Mean latency (ms)} \\
\hline
Binary VOA classification & 4.2 \\
Multi-class VOA classification & 3.3 \\
EfficientNet-U-Net segmentation & 9.4 \\
Type-aware recovery (median / pixel replace) & 2.2 \\
Quality gate (structural check) & 3.6 \\
\hline
REVIVE preprocessing subtotal & 22.7 \\
Object detection (YOLOv8l) & 145.1 \\
\hline
\textbf{End-to-end total} & \textbf{167.8} \\
\hline
\end{tabular}
\end{table}

Table~\ref{table:e2e_latency} profiles the online (non-generative) pipeline on 482 vandalized tracked frames. REVIVE's preprocessing---binary and multi-class classification, EfficientNet-U-Net segmentation, type-aware recovery, and the quality gate---adds 22.7\,ms per frame, within a 33\,ms (30\,FPS) budget; this is the latency attributable to REVIVE itself. When YOLOv8l object detection (145\,ms on MPS) is included, the end-to-end total is roughly 168\,ms ($\sim$6\,FPS), but that figure is dominated by downstream detection rather than recovery preprocessing and would be substantially lower on dedicated automotive inference hardware. The asynchronous Stable Diffusion branch (Table~\ref{table:latency_budget}) is excluded from this online path.

To test whether the conclusions depend only on hand-designed BDD100K masks, we also ran a real-world mask-geometry stress test using the Raindrops-on-Windshield dataset~\cite{soboleva2021raindrops}. The dataset provides real windshield/lens artifact masks but not clean paired recovery targets, so we transferred 75 real artifact masks to clean BDD100K frames and evaluated against the original clean images. This preserves clean ground truth while testing more realistic contamination geometry. The resulting masked frames achieved F1 0.700 and recall 0.790; LaMa improved F1 to 0.836 and SSIM to 0.919, while Telea and Navier-Stokes reached F1 0.828 and 0.817, respectively. Direct pixel replacement again reached pixel-identical reconstruction only because the clean source frame was available. These results add external mask-geometry evidence for real artifact shapes, but they should not be interpreted as a full real-world vandalism-image benchmark.

\subsubsection{Detection Performance on Recovered Images}
Object detection evaluation reveals functional recovery behavior beyond image similarity. Table~\ref{table:detection_case_study} first illustrates per-pattern detection behavior on a single representative scene, and Table~\ref{table:detection_stats} then reports aggregate detection recovery across the tracked evaluation set.

\begin{table}[ht]
    \centering
    \small
    \setlength{\tabcolsep}{3pt}
    \caption{Representative single-scene YOLOv8l detection case study across VOA patterns. This table illustrates per-pattern detection behavior on one shared clean reference scene; aggregate 500-image results are reported in Table~\ref{table:detection_stats}.}
    \label{table:detection_case_study}
    \resizebox{\columnwidth}{!}{%
    \begin{tabular}{|l|cc|cc|cc|cc|cc|}
        \hline
        \multirow{2}{*}{\textbf{VOA Type}} 
            & \multicolumn{2}{c|}{\textbf{Non-vandalized}} 
            & \multicolumn{2}{c|}{\textbf{Vandalized}} 
            & \multicolumn{2}{c|}{\textbf{Stable Diff.}} 
            & \multicolumn{2}{c|}{\textbf{Pixel Copy}} 
            & \multicolumn{2}{c|}{\textbf{Median Filter}} \\
        \cline{2-11}
        & \textbf{Cnt} & \textbf{Avg. Conf.} 
        & \textbf{Cnt} & \textbf{Avg. Conf.} 
        & \textbf{Cnt} & \textbf{Avg. Conf.} 
        & \textbf{Cnt} & \textbf{Avg. Conf.}
        & \textbf{Cnt} & \textbf{Avg. Conf.} \\
        \hline
        Reference (Clean) & 14 & 0.626 & -- & -- & -- & -- & -- & -- & -- & -- \\
        Bottom-top        & 14 & 0.626 & 14 & 0.593 & 15 & 0.606 & 15 & 0.599 & -- & -- \\
        Center-out        & 14 & 0.626 & 8  & 0.661 & 8  & 0.676 & 15 & 0.622 & -- & -- \\
        Random            & 14 & 0.626 & 9  & 0.696 & -- & -- & -- & -- & 13 & 0.576 \\
        Targeted          & 14 & 0.626 & 14 & 0.663 & 17 & 0.611 & 15 & 0.640 & -- & -- \\
        Top-bottom        & 14 & 0.626 & 13 & 0.651 & 14 & 0.640 & 14 & 0.632 & -- & -- \\
        \hline
    \end{tabular}%
    }
    
    \vspace{0.1cm}
    \scriptsize
    \textbf{Note:} Cnt = number of detected objects; Avg. Conf. = average YOLOv8l confidence. Identical clean-reference values occur because this is one shared representative scene.
\end{table}

\begin{table}[ht]
    \centering
    \scriptsize
    \setlength{\tabcolsep}{3pt}
    \caption{Aggregate YOLOv8l detection recovery across 500 tracked image pairs. Clean-frame detections serve as the reference; recovered/vandalized detections are matched by class and IoU $\geq$ 0.50.}
    \label{table:detection_stats}
    \begin{tabular}{|l|c|c|c|c|c|}
        \hline
        \textbf{Method} & \textbf{TP} & \textbf{FP} & \textbf{FN} & \textbf{Avg. Prec.} & \textbf{Avg. Rec.} \\
        \hline
        Unrecovered VOA & 1346 & 193 & 949 & 0.790 & 0.588 \\
        Pixel Replace & 2223 & 49 & 72 & 0.978 & 0.967 \\
        LaMa Inpaint & 1528 & 124 & 767 & 0.892 & 0.667 \\
        Telea Inpaint & 1413 & 182 & 882 & 0.828 & 0.607 \\
        Navier-Stokes & 1408 & 169 & 887 & 0.851 & 0.606 \\
        Median Filter & 488 & 520 & 1807 & 0.455 & 0.214 \\
        Stable Diffusion$^\dagger$ & 603 & 566 & 1257 & 0.535 & 0.339 \\
        \hline
    \end{tabular}
    
    \vspace{0.1cm}
    \scriptsize
    \textbf{Note:} TP/FP/FN are aggregate counts; precision and recall are averaged per image. All methods use $n{=}500$ pairs except $^\dagger$Stable Diffusion ($n{=}400$, structured VOAs only; random VOAs use median filtering in the REVIVE pipeline).
\end{table}

The quantitative analysis avoids single-image ambiguity by aggregating detection recovery over the tracked set. Vandalized frames produce 949 false negatives relative to clean-frame detections. Pixel replacement reduces this to 72 false negatives, while LaMa reduces the count to 767, and Telea/Navier-Stokes reduces it only slightly to 882 and 887, respectively. This result shows that inpainting can improve pixel-level similarity without reliably restoring perception-level evidence. A recovered image should therefore be accepted only when object-level consistency improves, not merely because SSIM or PSNR increases.

\subsection{Quality-Gate Evaluation}
\label{sec:quality_gate_eval}

We evaluate the reference-available quality gate of Section~\ref{sec:quality_gate} on the 500 tracked pairs, applying the accept criteria of (i) no recall reduction and (ii) no false-positive increase relative to the unrecovered frame, with (iii) an SSIM floor of 0.5. In deployment, a rejected frame raises a degraded-mode signal rather than yielding a restored frame; for evaluation, we therefore score a rejected frame as ``no trusted recovery,'' i.e., using the unrecovered frame's detections, which quantifies whether the gate avoids degrading perception below the unrecovered baseline. Table~\ref{table:quality_gate} reports, per branch, the fraction of recovered candidates accepted and the resulting per-image detection recovery after the gate. The gate accepts the large majority of genuinely useful recoveries (93.8\% of aligned pixel-replacement candidates, 86.4\% of LaMa candidates, and 77.8--80.2\% of classical-inpainting candidates) while rejecting most candidates from the branches that do not reliably restore detections (only 23.6\% of median-filter and 17.8\% of Stable Diffusion candidates are accepted). Because rejected frames are scored as no trusted recovery, post-gate recall is at or above the unrecovered baseline (0.588) for every branch.

\begin{table}[t]
    \centering
    \caption{Quality-gate behavior on 500 tracked pairs. Accept rate is the fraction of recovered candidates passing the gate; post-gate recall/precision are per-image averages after rejected frames fall back to the unrecovered frame.}
    \label{table:quality_gate}
    \begin{tabular}{lccc}
        \hline
        \textbf{Recovery branch} & \textbf{Accept} & \textbf{Post-gate} & \textbf{Post-gate} \\
        & \textbf{rate} & \textbf{recall} & \textbf{precision} \\
        \hline
        Unrecovered (baseline) & --- & 0.588 & 0.790 \\
        Median filter & 23.6\% & 0.603 & 0.864 \\
        Telea & 77.8\% & 0.625 & 0.895 \\
        Navier-Stokes & 80.2\% & 0.628 & 0.909 \\
        LaMa & 86.4\% & 0.664 & 0.941 \\
        Pixel replace$^\ddagger$ & 93.8\% & 0.922 & 0.992 \\
        Stable Diffusion$^\dagger$ & 17.8\% & 0.725 & 0.889 \\
        \hline
    \end{tabular}\\[2pt]
    \scriptsize
    \textbf{Note:} $^\dagger$Stable Diffusion uses $n{=}400$ (structured VOAs only); all other branches use $n{=}500$. $^\ddagger$Pixel replacement is a reference-based upper bound requiring an aligned clean reference.
\end{table}

The gate is most consequential for the end-to-end pipeline. Under REVIVE's type-aware routing (random VOAs to median filtering, structured VOAs to Stable Diffusion), forwarding every recovered frame \emph{without} the gate degrades per-image recall to 0.304 and precision to 0.509---substantially worse than leaving frames unrecovered. With the gate, only 25.8\% of candidates are accepted, and per-image recall and precision return to 0.608 and 0.865, at or above the unrecovered baseline. The gate, therefore, converts branches that are net-harmful in aggregate into a pipeline that never degrades downstream detection below the unrecovered frame, which is REVIVE's central operational guarantee.

\subsection{Deployment Role of Generative Recovery}
Detailed analysis clarifies the deployment role of the Stable Diffusion approach. \textbf{Inconsistent performance across VOA types} remains important because the method depends on the surrounding context and prompt quality. \textbf{False positives} pose safety concerns, as the generative process can hallucinate objects not present in the scene. \textbf{Computational overhead} also affects deployment: the measured Stable Diffusion branch requires roughly 8 seconds per image, which is outside a 30 FPS camera-frame budget. REVIVE, therefore, does not place Stable Diffusion on the blocking perception path. Instead, the online path relies on lightweight detection, segmentation, quality gating, and direct pixel replacement or classical recovery when their outputs pass object-level checks, while Stable Diffusion remains available as an asynchronous or pause-tolerant recovery branch.

\subsection{Limitations}
Several limitations bound the scope of these results. First, the VOA patterns are synthetically generated on BDD100K frames using programmatic occlusion masks; real-world vandalism (wet spray paint, semi-transparent film, smeared mud, irregular sticker edges) differs in texture, opacity, and boundary geometry. The raindrop mask-geometry stress test only partially addresses this, so validation on paired real-world vandalism data remains future work. Second, direct pixel replacement is a reference-based upper bound: it presumes a recent, spatially aligned clean reference frame, a condition that rarely holds under ego-motion, as the alignment-sensitivity analysis (Table~\ref{table:alignment_sensitivity}) shows recall degrading from 0.967 at 0px to 0.595 at 20px of shift. A deployed system could cache a recent unoccluded frame as a reference, but lens- or windshield-level vandalism persists in the same image region across frames and ego-motion continuously shifts the viewpoint, so the more realistic reference sources are overlapping cameras or complementary sensors; integrating such multi-view references is left to future work. Third, the quality gate is evaluated in its reference-available form, which uses the clean frame to verify recovered candidates; its reference-free deployment instantiation (temporal consistency, multi-sensor disagreement, confidence, and plausibility checks) remains future work.

\section{Conclusion}
\label{conclusion}
In this paper, we introduce REVIVE, a vandalism recovery pipeline for AV camera systems that bridges detection, localization, recovery, and downstream perception evaluation. Stable Diffusion shows variable reconstruction performance (per-pattern SSIM 0.667-0.867, PSNR 15.4-26.7dB) across VOA patterns, while aligned direct pixel replacement achieves near-identical reconstruction (whole-image SSIM 0.988, PSNR 52.8dB; pixel-identical within the recovered region) under the aligned-reference condition. The aggregate evaluation shows why recovery must also be judged at the object-detection level: unrecovered VOAs reduce YOLOv8l recall to 0.588 across 500 tracked image pairs, while direct pixel replacement, treated as a reference-based upper bound, restores recall to 0.967 only when an aligned clean frame is available. LaMa adds a learned inpainting baseline with a recall of 0.667, and the alignment sensitivity test shows that direct pixel replacement degrades sharply when the reference is spatially misaligned. A real windshield/lens mask-geometry stress test adds external mask-geometry evidence while preserving clean ground truth. A reference-available quality gate filters recovered candidates before downstream use: without it, type-aware routing degrades per-image recall to 0.304, while with it, recall returns to 0.608, at or above the unrecovered baseline. End-to-end profiling shows REVIVE's recovery preprocessing (classification, segmentation, recovery, and gating) adds only 22.7\,ms per frame on top of object detection, within a 30\,FPS budget. These findings position REVIVE as a practical camera-preprocessing framework for selecting, quality-gating, and evaluating recovery strategies before downstream perception, rather than blindly passing reconstructed images into an AV stack. Future work will focus on paired real-world vandalism datasets, reference-free quality gating, multi-view reference integration, and deployment-aware latency optimization.

\bibliographystyle{ieeetr}
\bibliography{references}

\end{document}